\documentclass[lettersize,journal]{IEEEtran}

\ifCLASSINFOpdf
\else
\fi
\newcommand{\hesam}[1]{{\color{blue} #1}}

\newcommand{\pfourmite}{\textit{P4Mite}}

\ifCLASSOPTIONcompsoc
  \usepackage[nocompress]{cite}
\else
  \usepackage{cite}
\fi

\usepackage{xurl}

\usepackage{graphics}
\usepackage{amsmath}
\usepackage{listings}
\usepackage{graphicx}
\usepackage{makecell}
\usepackage{support-caption}
\usepackage{subcaption}
\usepackage{lipsum}

\usepackage[table,xcdraw]{xcolor}
\usepackage{tikz}
\usepackage{comment}
\usepackage{amssymb}
\usepackage{pifont}
\usepackage{adjustbox} 
\usepackage{booktabs}  
  
\usepackage[table]{xcolor}
\usepackage[most]{tcolorbox}
\definecolor{main}{HTML}{757677}    
\definecolor{sub}{HTML}{DEE0E3}     
\newtcolorbox{boxH}{
    colback = sub, 
    colframe = main, 
    boxrule = 0pt, 
    leftrule = 6pt 
}

\begin{document}
%
\title{Performance Analysis of Decentralized Federated Learning Deployments}
%
%
%
%
\author{
	\IEEEauthorblockN{
		Chengyan Jiang\IEEEauthorrefmark{1},
		Jiamin Fan\IEEEauthorrefmark{2},
        Talal Halabi\IEEEauthorrefmark{3},
		Israat Haque\IEEEauthorrefmark{1}}

	\IEEEauthorblockA{
\IEEEauthorrefmark{1}\textit{Dalhousie University, Canada},
  \IEEEauthorrefmark{3}\textit{University of Victoria, Canada},
  \IEEEauthorrefmark{2}\textit{Laval University, Canada}
}}

\maketitle


\begin{abstract}

The widespread adoption of smartphones and smart wearable devices has led to the widespread use of Centralized Federated Learning (CFL) for training powerful machine learning models while preserving data privacy. However, CFL faces limitations due to its overreliance on a central server, which impacts latency and system robustness. Decentralized Federated Learning (DFL) is introduced to address these challenges. It facilitates direct collaboration among participating devices without relying on a central server. Each device can independently connect with other devices and share model parameters. This work explores crucial factors influencing the convergence and generalization capacity of DFL models, emphasizing network topologies, non-IID data distribution, and training strategies. We first derive the convergence rate of different DFL model deployment strategies. Then, we comprehensively analyze various network topologies (e.g., linear, ring, star, and mesh) with different degrees of non-IID data and evaluate them over widely adopted machine learning models (e.g., classical, deep neural networks, and Large Language Models) and real-world datasets.  The results reveal that models converge to the optimal one for IID data. However, the convergence rate is inversely proportional to the degree of non-IID data distribution. Our findings will serve as valuable guidelines for designing effective DFL model deployments in practical applications.

\end{abstract}

%
\IEEEpeerreviewmaketitle

\section{Introduction}

The advent of the Internet of Things (IoT), their applications (e.g., connected autonomous vehicles (CAVs), industrial IoT), and their volume of network and application data open up opportunities for developing data-driven decision making. However, collecting this data at cloud servers for further processing brings challenges like adequate transmission bandwidth and acceptable response time to end users. Edge computing is a paradigm of processing users' requests closer to them, e.g., in edge servers, to reduce resource demands while improving response time. Furthermore, federated learning emerged to offer data privacy with reduced bandwidth demand \cite{mcmahan2017communication}. Specifically, end devices do not share their data over the network to a central server to process them. Instead, the server shares the initial set of parameters with end devices to train a model on their local data. After completing the training, the devices share their trained parameters with the server to aggregate these parameters. The server shares the updated model parameters again, and the process continues until the model is globally trained with the coordination of the server. This methodology optimizes computational resources and accelerates developing and deploying intelligent network functions and services while preserving data privacy \cite{Federated_learning_for_internet_of_things:_A_comprehensive_survey}.

The above server-coordinated approach faces limitations like single points of failure, communication bottlenecks, and trust issues. To overcome these issues, Decentralized Federated Learning (DFL) facilitates the direct collaboration among devices, eliminating the need for a central server and enhancing robustness, trust, and resource distributions \cite{review_paper1,review_paper2}. Each device performs both computation and communication to serve as a server or a client. The decentralized approach enhances system fault tolerance and robustness and better protects data privacy by avoiding the security risks associated with centralized data storage. 

The effectiveness and reliability of DFL systems hinge upon critical factors such as network topology, non-IID data distributions, and training strategies \cite{review_paper2,review_paper1}. The standard topologies include linear, ring, and star, each having its strength in resource usage and model convergence rates. Non-IID data distributions refer to scenarios with devices having data with different distributions, characteristics, and statistical properties. Finally, DFL training strategies can be \textit{continuous} and \textit{aggregation} \cite{review_paper1}. In the former, the model parameters flow from one device to another in a sequential fashion, while the latter scheme deploys periodic aggregation of model parameters from multiple participating devices. Thus, depending on the application, users may choose a combination of network topology, degree of non-IID data distribution, and training strategy to meet their demand.  

Therefore, a thorough analysis is essential to understand how variations in network topology, degree of non-IID data, and different training strategies impact DFL performance. While existing studies have addressed some aspects of these challenges, notable gaps persist. Sheller \textit{et al.} \cite{sheller2020federated} compare convergence performance and accuracy across devices using continual training strategies with linear and ring topologies in DFL. Similarly, Chen \textit{et al.} \cite{chen2022decentralized} introduce a novel DFL solution utilizing a mesh topology and aggregate training strategy. However, neither study investigated the impact of the degree of non-IID data distribution, a crucial consideration in real-world applications. Furthermore, in \cite{coverage_CFL}, the authors focus on the convergence of Centralized Federated Learning (CFL) under varying degrees of non-IID data distributions using convex local models, leaving unexplored the efficacy of non-convex models. These gaps highlight the need for a comprehensive assessment to bridge the understanding of how DFL systems perform under diverse conditions, including both convex and non-convex models.

This paper rigorously analyzes the performance of DFL over the three factors mentioned above to train classical machine learning models (e.g., SVM, logic regression), neural networks (e.g., ResNet, DistilBERT), and large language models (LLMs) (e.g., MiniGPT-4) to shed light on their usage in emerging IoT applications. Specifically, we consider the combination of standard topologies with DFL training strategies along with IID and non-IID data distributions. First, we conduct a theoretical analysis of the convergence and generalization capabilities of convex models. Then, we extensively evaluate their performance to show alignment with the theoretical analysis. The results show that as the degree of non-IID increases, the actual loss value diverges from the ideal one, indicating that non-IID data distribution significantly affects the convergence of the DFL model. Furthermore, we evaluate non-convex models (ResNet, DistilBERT, and MiniGPT-4) to show their usage in DFL-based deployments. Our main contribution involves:

\begin{itemize}
    \item This work provides a comprehensive analysis of the factors influencing the convergence efficiency and generalization capacity of DFL models, covering both theoretical foundations and practical considerations.

    \item We rigorously establish the convergence of different network structures in DFL through convex optimization. This theoretical insight provides valuable guidance for researchers aiming to structure their DFL models effectively.
    
    \item To analyze the convergence performance of different model configurations in DFL under non-IID data, we employ various setups, including Continuous Linear, Continuous Ring, Aggregate Ring, Aggregate Linear, Aggregate Mesh, and Aggregate Star. The degree of non-IID data is quantified using a label imbalance method.

    \item We implement and assess five models: two convex (SVM and Logistic Regression) and three non-convex (ResNet, DistilBERT, and MiniGPT-4). Evaluation results show that all models across different topologies and training strategies converge under IID data. In the case of non-IID data, the performance is inversely proportional to the degree of distribution.
\end{itemize}

The remainder of the paper is structured as follows. Section \ref{sec:background-bg} describes some background concepts. Section \ref{sec:related} discusses related work. Section \ref{sec:convergence-rate} presents our convergence rate analysis. Section \ref{sec:implementation} describes our implementation details. Section \ref{sec:baseline} presents the evaluation setup. Section \ref{sec:evaluation result} discussed the evaluation results. Finally, Section \ref{section: Conclusion} concludes the paper.

\section{Background Concepts} \label{sec:background-bg}

\begin{figure}
    \centering
    \includegraphics[width=0.7\linewidth]{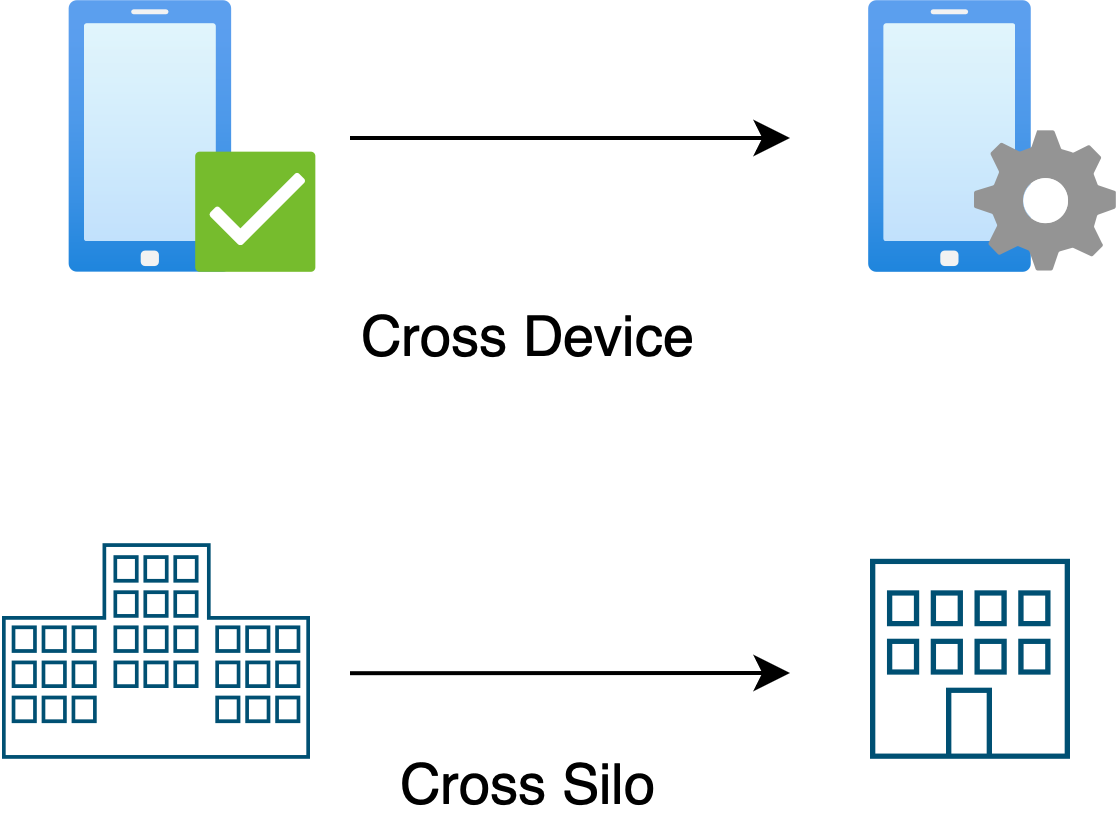}
    \caption{Cross silo vs Cross device}
    \label{fig:Cross silo vs Cross device}
\end{figure}
This section presents the necessary background information to follow the proposed work.

In Decentralized Federated Learning (DFL), model updates and parameter aggregation are carried out through direct communication among participating devices without having any central control. Thus, each device independently trains a model on its local data and exchanges model updates with other devices. They form different network topologies, e.g., linear, ring, star, and mesh \cite{review_paper1}. DFL excels in enabling the secure sharing of privacy-preserving data between real and virtual entities. This capability proves particularly advantageous in fostering task coordination and asynchronous knowledge exchange within domains such as robotics, energy, and utility sectors, with significant applications in Industry 4.0 and mobile services \cite{review_paper2}.  As shown in Fig.~\ref{fig:Cross silo vs Cross device}, there are two types of DFL: \textit{cross-silo} and \textit{cross-device} \cite{review_paper1,review_paper2}. The former enables distributed learning among organizations or data centers, i.e., a relatively small number of participating nodes. On the other hand, cross-device DFL involves a larger number of nodes, each with a modest amount of data and limited computational power. Our analysis applies to both cross-silo and cross-device DFL, i.e., to scenarios with varying numbers of devices .

\textbf{ Network Topology:} 
Participating devices in DFL training usually form linear, ring, star, and mesh network topologies, each with unique strengths. For instance, in ring and linear topologies, devices pass model parameters to downstream devices after finishing their turn, i.e., enable sequential parameter transmission without significant bandwidth demands. In the case of the star topology, the central device is responsible for parameter transmission and aggregation among all devices \cite{Optimization_Meet_DFL}. Note that unlike CFL, where the central device is a server, the central device in star topology is a participating node (server or client) with additional responsibilities of parameter processing. Thus, users may select a device with the most robust communication capabilities and the highest computational power as a central node. All devices in mesh topology are interconnected to offer the highest level of reliability in the presence of communication failures, where devices directly exchange parameters. 

The participating nodes in DFL can be categorized as a trainer, aggregator, proxy, and idle \cite{review_paper2}. The \textit{trainer and aggregator} nodes process data to train a model locally and aggregate parameters, respectively. The \textit{proxy} aggregates and forwards data between devices. Finally, \textit{idle} nodes are the ones who temporarily refrain from participating in the current training, e.g., maybe due to resource limitations. Thus, each node acts as a trainer in the continuous ring and linear topologies. Nodes are both trainer and aggregator in the aggregate ring, aggregate linear, and mesh topologies. Finally, every node except the central one serves as a trainer in the star topology, where the central node serves as the trainer and the aggregator. We do not consider proxy or idle nodes in our evaluation, as our focus is to assess the impact of topologies and training strategies on the performance of DFL.   

\textbf{Training Strategy:} 
DFL has two training strategies: \textit{aggregate} and \textit{continuous}. The former is similar to FedAvg \cite{first_FL}, i.e., participating devices collect parameters from others and aggregate them along with their own parameters. In a continuous scheme, participating devices start training from one of them and then sequentially pass parameters to subsequent ones to update their models. This scheme focuses on continuously learning and adapting to new data considering previously learned information, i.e., useful in scenarios where data distribution changes over time. 
Thus, if we combine the standard four topologies with training strategies, we get six deployment options for DFL training. 
Note that we exclude continuous star and continuous mesh as such combinations are not legitimate due to the ordering on message exchanges. Thus, we analyze the performance of these six deployments (Continuous Linear, Continuous Ring, Aggregate Linear, Aggregate Ring, Aggregate Star, and Aggregate Mesh) both mathematically and experimentally to show their applicability in different applications. 

\textbf{Non-IID Data:}
Non-IID data refers to the uneven or random statistical distribution of data on different clients. In federated learning, this means that the data on each device may vary significantly, which can adversely impact the distributed training process \cite{NON-IID_apact_accucry,FL_on_non-IID_survey}. 
Zhao \textit{et al.} \cite{NON-IID_apact_accucry} have demonstrated that the accuracy of FedAvg \cite{first_FL}, a fundamental algorithm in federated learning that averages the model parameters from each client weighted by the number of samples on the device to form a global model, diminishes notably with increasing data heterogeneity. Hangyu \textit{et al.} \cite{FL_on_non-IID_survey} have showcased various types of non-IID data in real-life scenarios, categorized into \textit{attribute} skew and \textit{label} skew. The former often associated with Horizontal Federated Learning (HFL), involves variations in the number of attributes on different devices. Notably, our study predominantly explores label skew due to its prevalence in Vertical Federated Learning (VFL), which is more relevant to our study on DFL. Label skew refers to differences in the number of labels for each device. In our investigation, we quantify the degree of non-IID Data in Section~\ref{sec:evaluation result}. 

In this study, we focus on Horizontal Federated Learning (HFL), the most common form of federated learning, which finds widespread applications in real-world scenarios, such as user device behavior data and regional medical records. By evaluating the performance of DFL under different network topologies within an HFL context, our research ensures broader applicability and practical relevance. Additionally, we emphasize the importance of Label Skew, a prevalent form of non-IID data in HFL, where clients' datasets often exhibit imbalanced label distributions. This imbalance can cause local models to overfit specific categories, thereby degrading the generalization capability of the global model. Label skew is widely observed in practice and significantly impacts the convergence and parameter aggregation of DFL. Investigating how DFL performs under different topologies with label skew enhances our understanding of optimizing decentralized federated systems, improving robustness and efficiency in non-IID data scenarios.

\textbf{Strongly Convex and \(L\)-smooth:}
A general assumption in federated learning-based training is that the models are strongly convex \cite{coverage_CFL}, which facilitates the stability and convergence of algorithms. This assumption ensures that the function has a unique global minimum without being trapped in local minima when searching for the optimal solution, thereby enhancing the stability and reliability of the algorithm. 
A function is \textit{strongly convex} if there is a constant \( \mu > 0 \) such that for all points \(x\) and \(y\) in its domain, the function lies above a quadratic curve with curvature \( \mu \). This property ensures the function has a unique global minimum, leading to faster convergence rates \cite{Uvarov1988}. A function is called \textit{L-smooth} if its gradient is L-Lipschitz continuous, i.e., the rate of change of the function is constrained by \(L\); in other words, the change in the gradient is bounded by \(L\) times the distance between \(x\) and \(y\). This property ensures that optimal learning does not overshoot the minimum. Both functions guide the convergence rate derivation process to determine the convergence speed; thus, they are incorporated into our mathematical analysis.

\section{Related Works} \label{sec:related}

This section presents the analysis of CFL and DFL similar to our study in order to position our contribution to the field. 

\textbf{Centralized Federated Learning:} 
Hangyu \textit{et al.} \cite{FL_on_non-IID_survey} evaluate various degrees of non-IID data in real-world scenarios and show that the degree of such data significantly impacts the prediction accuracy of the global model. Similarly, Zhao \textit{et al.} \cite{NON-IID_apact_accucry} show as data heterogeneity increases, the accuracy of the FedAvg algorithm decreases significantly. Their findings show that when data distribution differs greatly among different clients, the effectiveness of centralized federated learning is significantly compromised. Li \textit{et al.} \cite{coverage_CFL} mathematically prove the above claim and provide a solid foundation for understanding the impact of non-IID data on the convergence of federated learning. However, their work lacks analysis of different network topologies and only focuses on the FedAvg training strategy without exploring other training strategies. 

\textbf{Decentralized Federated Learning:}
Sheller \textit{et al.} \cite{sheller2020federated} are the first to analyze the performance comparison between continuous linear, continuous ring, and centralized federated learning. This study offers valuable insights into the impact of different topologies on DFL performance. However, the analysis lacks other standard topologies like star and mesh, the effect of non-IID data, and theoretical analysis. A similar study is performed in \cite{ICPP} without analyzing topological impact. Angelia \textit{et al.} \cite{Network_topology_and_communication-computation_tradeoffs_in_decentralized_optimization} analyze the network topologies of DFL using undirected and directed graphs. They apply the Metropolis algorithm to update directed graphs and the Push-sum algorithm for undirected graphs, analyzing the computational complexity and convergence speed of each method. However, the study does not show the impact of non-IID data over various network topologies. 

In summary, although existing studies have made some progress in exploring DFL, there are still unresolved issues. Particularly in handling non-IID data and complex network topologies, the performance and convergence of DFL require further analysis. We explore the performance of DFL through both theoretical~\ref{sec:convergence-rate} and experimental~\ref{sec:evaluation result} analysis. We first define six different DFL configurations based on two different training strategies and network topologies and conduct convergence analysis on them~\ref{sec:convergence-rate}. Then, we define the required non-IID settings in the experiments~\ref{sec:evaluation result} and measure the performance of the six different DFL configurations under varying degrees of non-IID data.

\section{Convergence Rate Analysis} \label{sec:convergence-rate}

This section develops optimization problems to derive the convergence rate of six different DFL deployments to shed light on their expected behavior. We use the notations in Table~\ref{table:notation_for_proof} throughout this section.  

\begin{table}[b]
\centering
\caption{Parameters used in our analysis}
\begin{tabular}{|c|l|}
\hline
\textbf{Notation} & \textbf{Description} \\ \hline
$\mathcal{L}oss(\cdot;\cdot)$ & User-specified loss function \\ \hline
$p_k$ & Weights of the \(k\)-th device \\ \hline

$N$ & Number of devices \\ \hline
$F(w)$ & Function for Decentralized Federated Learning (DFL) \\ \hline
$F^{\ast}$ & The optimal \(F\) under ideal conditions \\ \hline
$F_k$ & Objective for the \(k\)-th device \\ \hline
$F_k^{\ast}$ & The optimal \(F_k\) under ideal conditions \\ \hline
$\xi_k$ & SGD randomly chosen sample from \(k\)-th device \\ \hline
$w_k$ & Weights for \(k\)-th device \\ \hline
$n_k$ & Number of samples associated with the \(k\)-th device \\ \hline
$L$ & Parameter for L-smoothness \\ \hline
$\mu$ & Parameter for $\mu$-strong convexity \\ \hline
$Z$ & The level for the NON-IID\\ \hline
$NC$ & No coverage\\
\hline
\end{tabular}
\label{table:notation_for_proof}
\end{table}


\subsection{Problem Formulation} 

We consider a homogeneous environment (e.g., industrial edge nodes perform similar tasks) composed of devices with the same communication and computing capabilities. We do not distinguish between wired and wireless communication; instead, we assume an abstract mode of communication with a focus on the effect of network topologies on DFL performance. We develop distributed optimization problems for DFL deployment as shown in Equation~\ref{eq:DFL gobal objective function}, where the function focuses on optimizing the global model parameters by minimizing the global loss function \( F(w) \). Different network topologies determine how information is transmitted between devices with the same objective. This objective is to collaboratively compute local loss functions and share information to jointly optimize the global model.   


\begin{equation}
\label{eq:DFL gobal objective function}
\min_{w} \left\{ F(w) = \sum_{k=1}^{N} p_k F_k(w) \right\}
\end{equation}
Here, \( p_k \) denotes the weights of the \( k \)-th device (\( p_k \geq 0 \), \( \displaystyle \sum_{k=1}^{N} p_k = 1 \)), \( N \) is the total number of devices, and \( w \) represents the model parameters. The objective function for each device \( k \) (where \( 1 \leq k \leq N \)) using the entire dataset \(\mathbf{x}_k = \{x_{k,1}, \ldots, x_{k,n_k}\}\) is expressed as:

\begin{equation}
\label{eq:DFL_objective_function}
F_k(w) = \frac{1}{n_k} \sum_{j=1}^{n_k} \mathcal{L}oss(w; x_{k,j})
\end{equation}
where \(\mathcal{L}oss(\cdot; \cdot)\) represents a user-specified loss function and \( n_k \) represents the number of samples associated with the \( k \)-th device. If we uniformly select $\xi_{k}$ samples from device $k$'s dataset $\mathbf{x}_k$, the objective function becomes:
\begin{equation}
F_k(w) = \frac{1}{n_k}  \mathcal{L}oss(w; \xi_{k} )
\end{equation}

We assume that \(F_1, \ldots, F_N\) are all L-smooth and \(\mu\)-strongly convex \cite{coverage_CFL}, ensuring that the gradient does not change too rapidly and the value of the function between any two points is not only above the tangent line but also above the tangent line plus a positive term proportional to \(\|w - w'\|^2\). For \(k = 1, \ldots, N\), we also assume:
\[
E \|\nabla F_k(w_k, \xi_k) - \nabla F_k(w_k)\|^2 \leq \sigma_k^2
\]
and
\[
E \|\nabla F_k(w_k, \xi_k)\|^2 \leq G^2.
\]
These assumptions indicate that the noise's impact on gradient estimation is limited. We consider \(F_k^*\) to be the local optimal solution, i.e., it represents the minimum values of $F_k$. We use the term 
\[
F^{\ast} = \frac{1}{N} \sum_{k=1}^{N} F_k^*
\]
and 
\[
Z = F^{\ast}-F^{\ast}_k
\]
to quantify the degree of non-IID. If the data is IID, \(Z = 0\).

\subsection{Convergence Analysis}


We show that six different DFL deployments converge to the global optimal for strongly convex functions on Non-IID data. This means we are interested in evaluating $E\left( F\left( \bar{w} \right)  \right)  -F\left( w^{\ast }\right)$. Given that \( F_k \) is \( L \)-smooth, we can derive the following inequality:

\begin{equation}
F(\bar{w} )\leq F(w^{*})+\langle \nabla F(w^{*}),\bar{w} -w^{*}\rangle +\frac{L}{2} \| \bar{w} -w^{*}\|^{2} 
\end{equation}

Due to the monotonicity of the expected value, we have:
\[ E(F(\bar{w} ))\leq E(F(w^{*}))+E\left( \langle \nabla F(w^{*}),\bar{w} -w^{*}\rangle \right)  +\frac{L}{2} E\| \bar{w} -w^{*}\|^{2}  \] Subtracting \( F_k^* \) from both sides yields:
\begin{equation}
\begin{split}
E(F(\bar{w} ))-F^{*}\leq E(F(w^{*}))-F^{*}+\\E\left( \langle \nabla F(w^{*}),\bar{w} -w^{*}\rangle \right)  +\frac{L}{2} E\| \bar{w} -w^{*}\|^{2}  
\end{split} 
\end{equation}

We have the function: $E\left( F\left( w^{\ast }\right)  \right)  -F^{\ast }=0$.
Strongly convex functions possess the property that their gradient is zero at the optimal point $w^*$. This is because the gradient vector $\nabla F(w)$ points toward the fastest growth of the function $F(w)$. At the optimal point (local or global), the function reaches its minimum value, and consequently, the gradient $\nabla F(w^*)$ is zero. This can be expressed mathematically as: \[  \nabla F(w^*) = 0\]
As a result, for any vector $w - w^*$, the inner product $\langle \nabla F(w^*), w - w^* \rangle$ is zero:  \[  \langle \nabla F(w^*), w - w^* \rangle = 0  \]
Hence we obtain the following:

\begin{equation}
\label{equ:DFL function after first Simplified}
E(F(\bar{w} ))-F\left( w^{\ast }\right)  \leq \frac{L}{2} E||\bar{w} -w^{\ast }||^{2}
\end{equation}

The inequality in Equation~\ref{equ:DFL function after first Simplified} indicates that in a DFL deployment, the expected difference in loss values between the obtained model parameters \(\bar{w}\) and the optimal model parameters \(w^*\), denoted as \(E(F(\bar{w}) - F(w^*))\), can be bounded by the expected squared distance between them, \(\frac{L}{2} E \| \bar{w} - w^* \|^2\). In other words, this inequality shows that the smaller the distance between the average model parameters and the optimal parameters, the smaller the error in the loss function values. Our subsequent steps involve simplifying \(\| \bar{w} - w^* \|^2\) based on different DFL deployments. Thus, we will derive different convergence formulas for each topology.

\subsubsection{Continuous Linear} 

The Equation (\ref{eq:c_linear_Expressions}) shows the transfer of parameters. In detail, the continuous linear DFL~\cite{sheller2020federated} is structured as a line, with each node (device) connected to the next. Each device \( k \) receives the model parameters from the previous device $k-1$ and updates the model using its local dataset $x_{k}$. Subsequently, it forwards the trained parameters along the line to the next one. This sequential and iterative training continues along the line, fostering collaboration among the clients to enhance the learning and performance of the entire system collectively. This process can be formulated as follows:

\begin{equation}
\label{eq:c_linear_Expressions}
w_k=w_{k-1}-\eta_{k} \nabla F_{k}\left( w_{k-1},\xi_{k} \right) (1 \leq k \leq N)
\end{equation}
Here, $w_k$ denotes the updated parameter for device $k$, $\eta_{k}$ is the learning rate of device $k$ and $\nabla F_{k}\left( w_{k-1},\xi_{k} \right)$ represents the gradient of the local objective function $F_{k}\left( w_{k-1},\xi_{k} \right)$ for device $k$ based on the previous device's weight $w_{k-1}$ and its local data set $x_{k}$.

We set $w_k$ as the parameter trained on device $k$. 
Since the local model on each device is the same, it can be seen as a big model, hence we suppose $A_k=\nabla F_{k}\left( w_{k-1},\xi_{k} \right),\bar{A_k} =\nabla F_{k}\left( w_{k-1}\right)$; therefore, $E(A_k)=\bar{A_k}$, $w_k=w_{k-1}-\eta_{k} A_k$. Thus, Equation(~\ref{equ:DFL function after first Simplified}) is equivalent to:

\begin{equation}
    \label{eq:Continous Linear step one}
    \begin{split}
    \begin{gathered}||w_k-w^{\ast }||^{2}=||w_{k-1}-\eta_{k} A_{k} -w^{\ast }-\eta_{k} \bar{A}_{k} +\eta_{k} \bar{A}_{k} ||^{2}\\=||w_{k-1}-\eta_{k} \bar{A}_{k} -w^{\ast }||^{2}+2\eta_{k} <w_{k-1}-w^{\ast }-\eta_{k} \bar{A}_{k} ,\\\bar{A}_{k} -A_{k}>+\eta^{2}_{k} ||A_{k}-\bar{A}_{k} ||^{2} \end{gathered} 
    \end{split}
\end{equation}
There are three parts in Equation (\ref{eq:Continous Linear step one}): $||w_{k-1}-\eta_{k} \bar{A}_{k} -w^{\ast }||^{2}$, $2\eta_{k} <w_{k-1}-w^{\ast }-\eta_{k} \bar{A}_{k} ,\bar{A}_{k} -A_{k}>$, and $\eta^{2}_{k} ||A_{k}-\bar{A}_{k} ||^{2}$. We can see that $2\eta_{k} <w_{k-1}-w^{\ast }-\eta_{k} \bar{A}_{k} ,\bar{A}_{k} -A_{k}>=0$ because when we want to get the expectation for $2\eta_{k} <w_{k-1}-w^{\ast }-\eta_{k} \bar{A}_{k} ,\bar{A}_{k} -A_{k}>$, the $E(A_{k})=\bar{A}_{k}$, which means $\bar{A}_{k} -A_{k}=0$. Therefore the dot product is also zero.

\begin{equation}
\label{eq:Continous Linear step two}
\begin{split}
    ||w_{k-1}-\eta_{k} \bar{A}_{k} -w^{\ast }||^{2}=||w_{k-1}-w^{\ast }||^{2}\\-2\eta_{k} <w_{k-1}-w^{\ast },\bar{A}_{k} >+\eta^{2}_{k} ||\bar{A}_{k} ||^{2}
\end{split}
\end{equation}
We can see there are three part in the equation (\ref{eq:Continous Linear step two}), since $\eta^{2}_{k} ||\bar{A}_{k} ||^{2}=\eta^{2}_{k} ||\nabla F_{k}\left( w_k\right)  ||^{2}$ and because all the local models are L-smooth, we can bound $||\bar{A}_{k} ||^{2}$ as:
\begin{equation}
\eta^{2}_{k} ||\nabla F_{k}\left( w_k\right)  ||^{2}\leq 2\eta^{2} L\left( F_{k}\left( w_k\right)  -F^{\ast }_{k}\right)
\end{equation}
Then, we do the bonding at $-2\eta_{k} <w_{k-1}-w^{\ast },\bar{A}_{k}>$, by the definition of the strongly convex: 
\begin{align}
-2\eta_{k} \langle w_{k-1}-w^{\ast }, \nabla F_{k}\left( w_{k-1}\right) \rangle &\leq -2\eta_{k} \left( F_{k}\left( w_{k-1}\right)  -F_{k}\left( w^{\ast }\right) \right. \nonumber \\
&\quad \left. -\frac{\mu }{2} \parallel w_{k-1}-w^{\ast }\parallel^{2}_{2} \right)
\end{align}

Then, we substitute the above two equations into Equation~\ref{eq:Continous Linear step two}:
\begin{equation}
\begin{split}
||w_{k-1}-\eta_{k} \bar{A}_{k} -w^{\ast }||^{2}\leq ||w_{k-1}-w^{\ast }||^{2} -2\eta_{k}(F_{k}\left( w_{k-1}\right)\\-F_{k}\left( w^{\ast }\right) -\frac{\mu }{2} \parallel w_{k-1}-w^{\ast }\parallel^{2}_{2} )+2\eta^{2}_{k} L\left( F_{k}\left( w_k\right)  -F^{\ast }_{k}\right)  
\end{split}
\end{equation}

\begin{equation}
    \label{eq:Continous Linear step three}
    \begin{split}
    ||w_{k-1}-\eta_{k} \bar{A}_{k} -w^{\ast }||^{2}\leq ||w_{k-1}-w^{\ast }||^{2}-2\eta_{k} [F_{k}\left( w_{k-1}\right)  \\-F_{k}\left( w^{\ast }\right)  -\frac{\mu }{2} \parallel w_{k-1}-w^{\ast }\parallel^{2} ]+(2\eta_{k} )^{2}L\left( F_{k}\left( w_k\right)  -F^{\ast }_{k}\right)   \\\leq ||w_{k-1}-w^{\ast }||^{2}-2\eta_{k} F_{k}\left( w_{k-1}\right)  +2\eta_{k} F_{k}\left( w^{\ast }\right)  \\+\mu \eta_{k} \parallel w_{k-1}-w^{\ast }\parallel^{2} +(2\eta_{k} )^{2}L\left( F_{k}\left( w_k\right)  -F^{\ast }_{k}\right) 
    \\ =\left( 1+\mu \eta_{k} \right)  ||w_{k-1}-w^{\ast }||^{2}-2\eta_{k} \left( F_{k}\left( w_{k-1}\right)  -F^{\ast }_{k}\right)  \\+(2\eta_{k} )^{2}L\left( F_{k}\left( w_k\right)  -F^{\ast }_{k}\right)  
    \end{split}
\end{equation}
Then, we let the $J=-2\eta_{k}\left( F_{k}\left( w_{k-1}\right)  -F^{\ast }_{k}\right)  +(2\eta_{k} )^{2}L\left( F_{k}\left( w_k\right)  -F^{\ast }_{k}\right))$.

Since $F^{\ast }_{k}-F_{k-1}\leq F^{\ast }-F_{k-1}\leq F^{\ast }-F^{\ast }_{k}$, therefore, 
$J\leq 2\eta_{t} \left( F^{\ast}-F^{\ast }_{k}\right)+2\eta^{2}_{t} L\left( F_{k}-F^{\ast }_{k}\right)$. 
To bound $( F_{k}-F^{\ast }_{k})$, since $F_{k}$ is $L$-smooth, we can get 
$\left( F_{k}-F^{\ast }_{k}\right)  \leq \frac{L}{2} \parallel w_k-W^{\ast }\parallel^{2} 
+\left( w_k-W^{\ast}\right)^{T}  \nabla F_{k}\left( W^{\ast }\right)$. 
Since $\nabla F_{k}\left( W^{\ast }\right)=0$, we have:
$J\leq 2\eta_{k} Z+\eta^{2}_{k} L^{2}\parallel w_k-w^{\ast }\parallel^{2} $, 
and Equation~\ref{eq:Continous Linear step three} becomes 
$\left( 1+\mu \eta_{k} \right)  ||w_{k-1}-w^{\ast }||^{2}+2\eta_{k} Z+\eta^{2}_{k} L^{2}\parallel w_k-w^{\ast }\parallel^{2}$. 
Therefore, Equation~\ref{eq:Continous Linear step one} becomes:

\begin{equation}
    \label{equation: 23}
    \begin{split}
       \parallel w_k-w^{\ast }\parallel^{2} \leq \left( 1+\mu \eta_{k} \right)  ||w_{k-1}-w^{\ast }||^{2}\\+2\eta_{k} Z+\eta^{2}_{k} L^{2}\parallel w_k-w^{\ast }\parallel^{2} +\eta^{2}_{t} \parallel A_{t}-\bar{A}_{t} \parallel^{2} 
    \end{split}
\end{equation}

By taking expectations on both sides of Equation~\ref{equation: 23} we get:
\begin{equation}
    \begin{split}
    E\parallel w_k-w^{\ast }\parallel^{2} \leq \left( 1+\mu \eta_{k} \right)  E||w_{k-1}-w^{\ast }||^{2}\\+2\eta_{k} Z+\eta^{2}_{k} L^{2}E\parallel w_k-w^{\ast }\parallel^{2} +\eta^{2}_{k} E\parallel A_{k}-\bar{A}_{k} \parallel^{2} 
    \end{split}
\end{equation}

We next bond the $\eta^{2}_{t} E\parallel A_{t}-\bar{A}_{t} \parallel^{2}$, by the assumption, $\eta^{2}_{t} E\parallel A_{t}-\bar{A}_{t} \parallel^{2} \leq \eta^{2}_{t} \sigma^{2}_{k} $.

Thus, we can see that in the final Function~\ref{eq: C_linear_fianl}, if the \( Z \) (degree of non-IID) approaches zero, the left part will be bounded by a constant. This means that when the data distribution on each client is the same, the DFL model will converge. However, as the level of non-IID increases, the value on the right side of the equation increases, causing the gap between the actual loss and the ideal loss to increase. This leads to the model becoming more divergent, as follows:

\begin{align}
\label{eq: C_linear_fianl}
 E\left[ F\left( w_k\right)  -F^{\ast }\right]  &\leq \frac{L}{2} \left[ \left( 1+\mu \eta_{k} +\eta^{2}_{k} L^{2}\right) E\|w^{0}-w^{\ast }\|^{2} \right. \nonumber \\
 &\quad \left. +2\eta_{k} Z + \eta^{2}_{k} \sigma^{2}_{k} \right]
\end{align}
where $w^{0}$ represents the initial parameters.

\subsubsection{Continuous Ring}

The continuous ring DFL~\cite{sheller2020federated} comprises a network of devices interconnected in the form of a ring. We want to mention that the continuous ring is very similar to continuous linear because they have the same training strategy.

We define $d$ as the cumulative number of times the model parameters are passed to all devices, which is an indicator of the flow of the model throughout the network, where $d\in 1,...,N\ast t$. Assuming that device$1$ received initial parameters $w^0$, in each training round $t$ ($t>0$), the $k$-th ($k=d-\left( t-1\right) N$) device updates the initial model parameters using its data and then transmits the updated parameters to the ($k+1$)-th device. 

The subsequent device considers the received model parameters from the previous device as the initial model and uses its data to update it. When the model is transmitted to the last device, that device sends the trained parameters back to the first device. Subsequently, the first device sets these parameters as the initial values and initiates training again. This cyclic process is mathematically represented by the following equation:

\begin{equation}
\begin{split}
\  w_{d}=w_{d-1}-\eta_{d} \nabla F_{d}\left( w_{d-1},\xi_{d} \right)  \\ k=d-\left( t-1\right)  N
\end{split}
\end{equation}

Here, $t$ represents the number of complete cycles from the first client to the last client and then back to the first client, constituting one full training cycle. Each complete cycle allows every device an opportunity to update and pass along the model parameters, considered as one complete training round.
Therefore, the coverage analysis of ring continuous is similar to the linear continuous:

\begin{align}\label{Continous Ring}
E\left[ F\left( w_{d} \right) -F^{\ast } \right] &\leq \frac{L}{2} \left[ \left( 1+\mu \eta_{d} +\eta^{2}_{d} L^{2} \right) E||w^{0}-w^{\ast }||^{2} \right. \nonumber \\
&\quad \left. +2\eta_{d} Z+\eta^{2}_{d} \sigma^{2}_{d} \right]
\end{align}

Since C\_ring and C\_linear use the same aggregation strategy, their mathematical convergence analysis formulas are similar. Therefore, the derivation process is omitted. By observing the formula, we can see that as \( Z \) (the degree of non-IID) increases, the value on the left side becomes larger, and the gap between the optimal solution and the actual value on the right side also increases, causing the model to become more divergent.

\subsubsection{Aggregate Linear}

Aggregate linear DFL refers to the use of a linear network topology, combined with an aggregation training strategy. Similar to the continuous linear, this deployment comprises a series of interconnected devices aligned linearly. Each device within this network possesses a unique dataset. After training on Device 1, it transmits its parameters to Device 2. 

Following the same procedure as the continuous linear approach, Device 2 trains based on the parameters from Device 1. Device 2 then transmits both its own parameters and those from Device 1 to Device 3. Device 3 aggregates the parameters from both Device 1 and Device 2, and then trains based on the aggregated parameters. The aggregation follows a method similar to FedAvg. 

Specifically,  we calculate the cumulative sample count of Devices 1 and 2, termed as $s_{sum}$. Additionally, we denote the individual sample counts for Device1 and Device2 as $s1$ and $s2$, respectively. The final parameter set transmitted from Device2 to Device3 is a weighted combination ($w1$ is the weight from client1 to client2, $w2$ is the weight after client2 trained its own dataset) defined as:
$\frac{s1}{s_{sum}} \times w_{1}+\frac{s2}{s_{sum}} \times w_{2}$.


This process is mathematically represented by the following equation; when k is at most 2:
\begin{equation}
w_k=w_{k-1}-\eta_{k} \nabla F_{k}\left( w_{k-1},\xi_{k} \right) 
\end{equation}

When k is larger than 2, we get:
\begin{equation}
\begin{split}
 w_k=\left( \frac{\left( s_{k-1}\right)  w_{k-1}+\left( s_{k-2}\right)  w_{k-2}}{\displaystyle \sum^{k-1}_{i=1} s_{i}} \right)  -\eta_{k} \nabla F_{k}\\\left( \left( \frac{\left( s_{k-1}\right)  w_{k-1}+\left( s_{k-2}\right)  w_{k-2}}{\displaystyle \sum^{k-1}_{i=1} s_{i}} \right)  ,\xi_{k} \right)     
\end{split}
\end{equation}

Next, we will analyze its convergence:
For the case where $w_k=M_{k}-\eta_{k} \nabla F_{k}\left( M_{k},\xi_{k} \right)$ and $ M_{k}=\left(\frac{s_{k-1}w_{k-1}+s_{k-2}w_{k-2}}{\displaystyle\sum^{k-1}_{i=k} s_{i}}  \right) $ we let $H=\nabla F_{k}\left( M_{k},\xi_{k} \right)$ and $\bar{H} =\nabla F_{k}\left( M_{k}\right) $ Thus, $E(H)=\bar{H}$ and $w_k=M-\eta_{k} H$, and we get:

\begin{equation}
\begin{gathered}\parallel w_k-w^{\ast }\parallel^{2} =\parallel M-\eta_{k} H-w^{\ast }-\eta_{k} \bar{H} +\eta_{k} \bar{H} \parallel^{2} \\ =\parallel M-\eta_{k} \bar{H} -w^{\ast }+\eta_{k} \bar{H} -\eta_{k} H\parallel^{2} \end{gathered} 
\end{equation}

The same applies to the continuous linear, Equation~\ref{equ:DFL function after first Simplified} is equivalent to
$| M-\eta_{k} \bar{H} -w^{\ast }\|^{2} -2<M-\eta_{k} \bar{H} -w^{\ast },\eta_{k} \bar{H} -\eta_{k} H>+\eta^{2}_{k} \parallel \bar{H} \parallel^{2}$
and we will bond the $| M-\eta_{k} \bar{H} -w^{\ast }\|^{2}$
 first. $\| M-\eta_{k} \bar{H} -w^{\ast }\|^{2}=\  \| M-w^{\ast }\|^{2} -2\eta_{k} <M-w^{\ast },\bar{H} >+\eta^{2}_{t} \| \bar{H} \|^{2} \\$ and we will bond the $2\eta^{2}_{k} <M-w^{\ast },\bar{H} >$ and $\eta^{2}_{k} \| \bar{H} \|^{2} $

\begin{equation}
\begin{split}
  2\eta_{k}^{2} \langle M - w^{\ast}, \bar{H} \rangle & \leq -2\eta_{k} \left( F_{k}(M) - F(w^{\ast}) \right. \\
  & \quad \left. + \frac{\mu}{2} \| M - w^{\ast} \|_{2}^{2} \right)
\end{split}
\end{equation}

\begin{equation}
    \eta^{2}_{k} \| \nabla F_{k}\left( M \right) \|^{2} \leq 2\eta^{2}_{k} L\left( F_{k}\left( M \right) -F^{\ast }_{k} \right) 
\end{equation}

Then, we get 
\begin{equation}
    \begin{split}
        \| M - \eta_{k} \bar{H} - w^{\ast} \|^{2} \leq & \left( 1 + \mu \eta_{k} \right) \| M - w^{\ast} \|^{2} \\
        & + 2 \eta_{k} \left( F_{k} \left( w^{\ast} \right) - F_{k} \left( M \right) \right) \\
        & + 2 \eta_{k}^{2} L \left( F_{k} \left( M \right) - F^{\ast}_{k} \right)
    \end{split}
\end{equation}

In the same way as the continuous linear, we can get the equation:
\begin{equation}
    \begin{split}
    \begin{gathered}A\leq \left( 1+\mu \eta_{k}  \right) \| M-w^{\ast }\|^{2} +2\eta_{k} Z+\eta^{2}_{k} L^{2}\| M-w^{\ast }\|^{2} \\ \| w_k-w^{\ast }\|^{2} \leq \left( 1+\mu \eta_{k}  \right) \| M-w^{\ast }\|^{2}\\ +2\eta_{k} Z+\eta^{2}_{k} L^{2}\| M-w^{\ast }\|^{2} +\eta^{2}_{k} \sigma^{2}_{k} \end{gathered} 
    \end{split}
\end{equation}
By the strong convex property, we can replace the $ M_{k}=\left( \frac{s_{k-1}w_{k-1}+s_{k-2}w_{k-2}}{\displaystyle \sum^{k-1}_{i=k} s_{i}}  \right) $ with  $V_{k}=\left( \frac{s_{k-1}w^{0}+s_{k-2}w^{0}}{\displaystyle\sum^{k-1}_{i=k} s_{i}}  \right) $

Thus, we get the function for aggregate linear:
\begin{equation}
    \begin{split}
        E(F(\bar{w} ))-F\left( w^{\ast } \right) )\leq \frac{L}{2} [\left( 1+\mu \eta_{k}  \right) \| V_{k}-w^{\ast }\|^{2} \\+2\eta_{k} Z+\eta^{2}_{k} L^{2}\| V_{k}-w^{\ast }\|^{2} +\eta^{2}_{k} \sigma^{2}_{k} ]
    \end{split}
\end{equation}

We see that all parameters except \( Z \) are constants, which means that the right side of the formula is only affected by \( Z \). However, an increase in \( Z \) will cause the right side to increase as a whole. This simultaneously affects the value of the left side of the formula, leading to poorer overall convergence of the model.

\subsubsection{Aggregate Ring}
 
The aggregate ring is similar to the aggregate linear because they have the same training strategy. A series of interconnected devices form a ring. Each device possesses a unique dataset. There is only one difference. When running to the last device, the last aggregated parameters will be sent back to the first device.

This process is mathematically represented by the following equation; when k is at most 2:
\begin{equation}
w_{d}=w_{d-1}-\eta_{d} \nabla F_{d}\left( w_{d-1},\xi_{d} \right)  
\end{equation}
when k is larger than 2:
\begin{equation}
\begin{split}
 w_{d}=\left( \frac{\left( s_{d-1}\right)  w_{d-1}+\left( s_{d-2}\right)  w_{d-2}}{\displaystyle \sum^{d-1}_{i=1} s_{i}} \right)  -\eta_{d}\\ \nabla F_{d}\left( \left( \frac{\left( s_{d-1}\right)  w_{d-1}+\left( s_{d-2}\right)  w_{d-2}}{\displaystyle \sum^{d-1}_{i=1} s_{i}} \right)  ,\xi_{d} \right) 
\\k=d-\left( t-1\right)  N
\end{split}
\end{equation}

And since the aggregate ring have the same training strategy, we get:
\begin{align}
E(F(\bar{w} ))-F\left( w^{\ast } \right) &\leq \frac{L}{2} \left[ \left( 1+\mu \eta_{d}  \right) \| V_{d}-w^{\ast }\|^{2} \right. \nonumber \\
&\quad \left. + 2\eta_{d} Z + \eta^{2}_{d} L^{2}\| V_{d}-w^{\ast }\|^{2} + \eta^{2}_{d} \sigma^{2}_{d} \right]
\end{align}

Since A\_ring and A\_linear use the same aggregation strategy, their mathematical convergence analysis formulas are similar. Therefore, the derivation process is omitted. By observing the formula, we can see that as \( Z \) (the degree of non-IID) increases, the value on the left side becomes larger, and the gap between the optimal solution and the actual value on the right side also increases, causing the model to become more divergent.

\subsubsection{Aggregate Star}

The aggregate star is similar to CFL\cite{first_FL,coverage_CFL}, however, the center device no longer only plays the role of a server, instead, it can participate in the training of local models. First, select a central device as the center of the whole deployment. After that, the initial parameters are issued by the central device to each device, and then each device (including the central device) is trained based on the initial parameters. After that, the central  device receives the parameters of all devices for aggregation. This process is  repeated iteratively. 



This process is mathematically represented by the following equation: 
\begin{equation}
w_k=\sum^{N}_{i=1} \left( \frac{s_{i}w_{i}}{\displaystyle\sum^{n}_{j=1} s_{j}} \right)  -\eta_{k} \Delta F_{k}\left( \sum^{N}_{i=1} \left( \frac{s_{i}w_{i}}{\displaystyle\sum^{n}_{j=1} s_{j}} \right)  ,\xi_{k} \right)   
\end{equation}

To analyze the convergence, the  aggregate star and mesh DFL are similar with CFL, Li et al. \cite{coverage_CFL} have proved it as: 
$\mathbb{E}[F(w_{T})] - F^* \leq \frac{2\kappa}{\gamma + T} \left( \frac{B}{\mu} + 2L \|w^0 - w^*\|^2 \right),
B = \displaystyle\sum_{k=1}^{N} p_k^2 \sigma_k^2 + 6LZ + 8(E - 1)^2 G^2$, where $T$ is the number of iterations.

We see that all parameters except \( Z \) are constants, which means that the right side of the formula is only affected by \( Z \). However, an increase in \( Z \) will cause the right side to increase as a whole. This simultaneously affects the value of the left side of the formula, leading to poorer overall convergence of the model.


\subsubsection{Aggregate Mesh}

Unlike the aggregation star, each device can play the role of the central device and can be trained for local models. First of all, each device has preset parameters. Each device is trained locally according to the parameters and then sent to other devices. Each device aggregates parameters separately, and the aggregated parameters are recorded as the initial parameters of the next round. The next round of training is carried out based on the initial parameters. The training process is the same as the aggregate star.

\begin{boxH}
\textbf{Key Takeaways:} 
In general, non-IID data distribution has adverse impact on the convergence rate irrespective of topologies and training strategies. However, different topologies converge differently. 
In continuous linear and ring-based DFL, the performance depends on \( Z \) (representing the degree of non-IID distribution). As \( Z \) increases, the convergence of the model worsens. When \( Z = 0 \), the gap between the optimal and the actual solution becomes a constant value, indicating that the model has converged. 
In aggregate linear and ring structures, the performance depends on two variables, \( V_k \) and \( Z \). Here, \( V_k \) represents the ratio of the sum of the samples of the current and the previous devices to the total number of samples.
If this ratio is 1, aggregate and continuous structures are equivalent; however, when the ratio is greater or less than 1, continuous structures converges better than the aggregate and vice versa, respectively. 
Finally, in star and mesh structures, performance depends not only on \( Z \) but also on \( T \) (the number of rounds), i.e., the greater the number of rounds \( T \), the better the convergence.
\end{boxH}

\section{Implementation Details} \label{sec:implementation}

This section presents our evaluation process, including implementation and evaluation details, model and data selection, and non-IID data generation.

We evaluate the performance of all models in DFL deployment over a NVIDIA Quadro RTX 8000 GPU. This device is equipped with 48GB of memory, driver version 535.86.05, and CUDA version 12.2. We simulate five different devices on this server, each capable of training models and sending parameters to each other. The models are implemented using Python 3.12.1, NumPy version 1.26.4, scikit-learn version 1.5.0 and PyTorch version 2.3.0+cu121. Finally, we implement MiniGPT-4 using Python 3.10 and PyTorch version 2.3.0+cu121 following \cite{minigpt-4}. Unlike traditional and deep learning models, due to the enormous size of MiniGPT-4, its training is divided into two parts: the first part trains the connections between the vision and language models, and the second part involves fine-tuning the model. In this experiment, we only trained the fine-tuning part. 

\textbf{Model and Data Selection:}
To ensure that our research adapts to multiple fields, we have chosen three different types of learning models: traditional, deep learning, and large language models. The loss functions of traditional models are often convex, while those for deep learning and large language models are nearly convex. The models and their datasets are listed in Table~\ref{table:model-data}.

\begin{table}[h]
\centering
\caption{Model \& Dataset Table}
\begin{tabular}{|c|l|}
\hline
\textbf{Model} & \textbf{Dataset} \\ \hline
SVM & Breast Cancer Dataset \\ \hline
Logistic Regression & Breast Cancer Dataset \\ \hline
ResNet-18 & MNIST dataset\\ \hline
DistilBERT  & TREC\_6 dataset \\ \hline
MiniGPT-4 &  cc sub dataset \\ \hline
\end{tabular}
\label{table:model-data}
\end{table}

We choose Support Vector Machines (SVM) \cite{svm} and Logistic Regression \cite{logistic_regression} as they are the most widely adopted baseline models both in classifications and regressions.
In the case of deep learning models, we select ResNet-18 \cite{resnet} for vision and DistilBERT \cite{bert} for NLP tasks. ResNet-18 has a relatively shallow structure, low computational cost, and is suitable for experiments in resource-constrained environments while still providing strong image feature extraction capabilities. Similarly, DistilBERT is a lighter variant of the BERT model suitable for edge deployments. Finally, we use MiniGPT-4 \cite{minigpt-4} as the LLM representative to test in the edge environment. We deploy Stochastic Gradient Descent (SGD) as the optimizer during the training of these chosen models.

\textbf{Non-IID Data Distribution:}
In our experiment, we consider the most prevalent form of NON-IID data ---label skew ---as this is commonly observed in vertical Federated Learning models \cite{FL_on_non-IID_survey}. Label skew is defined as the variation in the distribution of dataset labels across different clients. In the following, we present the process of generating multiple levels using approaches presented in \cite{FedCPD}.  

SVM and logistic regression perform binary classifications on the Breast Cancer Dataset, containing only two labels (true vs. false); thus, we employ three levels of non-IID data. To determine these levels, we used KL divergence as the metric \cite{FedCPD}. We first set the positive and negative distribution of data labels. For example, if \([0.1,0.3,0.5,0.7,0.9]\) and \([0.9,0.7,0.5,0.3,0.1]\) are the positive and negative labels across five devices, respectively, then there will be 10\% of the positive and 90\% of the negative labels from the complete dataset on the first device. Similarly, the second device will have 30\% positive labels and 70\% negative labels. 
We consider three different non-IID distributions:  \([0.5,0.6,0.7,0.8,0.9]\), and \([0.1,0.3,0.5,0.7,0.9]\),\([1,0,0.7,1,0]\). We compare these distributions with the complete data distribution (\([1,1,1,1,1]\)) and calculate the KL divergence, obtaining values of  0.0206,0.18013 and 0.52371 which represent three different levels. This method ensures that each device has all labels, but the label distribution varies between devices.

In the case of multi-label data used in deep learning, we consider two scenarios: 1) each device has all labels, but the label distribution differs between devices, and 2) each device does not have all labels, and the label distribution differs between devices. In total, our experiment has five different levels of non-IID settings. Levels 1 and 2 are the same as in the traditional models, whereas the remaining levels do not have the same levels. Specifically, devices in Level 3 have 90\% of the labels from the complete dataset, Level 4 has 70\% of the labels, and Level 5 has 50\% of the labels.  
The cc\_sub dataset is inherently a non-IID dataset composed of images and text and thus does not need any processing. Each image and its associated text in this dataset differ from others, naturally exhibiting non-IID characteristics \cite{minigpt-4}.

\textbf{Evaluation Scheme and Metrics:}
We have defined a baseline model, referred to as \(F^*\), representing the state with the minimum loss value. Mathematical analysis has confirmed that \(F^*\) constitutes the optimal solution under ideal conditions. Such perfect conditions presume the existence of a machine with unlimited computational power and memory capable of processing the entire dataset. In essence, the baseline can be regarded as the optimal model outcome obtained under ideal circumstances. However, the real deployment differs from the ideal one, i.e., we can measure how much a DFL-based model training differs from the baseline. 

Our first evaluation verifies whether all six DFL deployments support convergence for the chosen models. 
Thus, we set each device with the complete dataset and the same hyperparameters as the baseline model. In the second evaluation, we consider non-IID data across devices, i.e., the models operate with the same hyperparameter as the baseline models but with different degrees of non-IID datasets. The convergence is measured by examining loss curves. Specifically, when the loss curve becomes gradually flat or the training and validation loss curves intersect, we consider the models converged. We also measure the model accuracies as defined below.

We use binary classification for traditional machine learning datasets and multi-class classification for deep learning datasets. 
The binary classification F1 score is given by
\[
F1 = \frac{2 \cdot \text{Precision} \cdot \text{Recall}}{\text{Precision} + \text{Recall}},
\]
while the multi-class F1 score is calculated using macro-averaging as
\[
F1_{\text{macro}} = \frac{1}{N} \sum_{i=1}^{N} F1_i.
\]
This approach comprehensively measures the model's performance across different classes. For multi-class tasks, the macro-averaged F1 score reflects the balanced performance of each class, avoiding bias towards classes with larger data volumes.

To test the LLM accuracy as described in \cite{minigpt-4}, we ask the model four different questions in the form of multiple-choice questions: 
\begin{itemize}
    \item Help me draft a professional advertisement for this.
    \item Can you craft a beautiful poem about this image?
    \item Explain why this meme is funny.
    \item How should I make something like this?
\end{itemize}
We provide the model with an image and then ask these questions to see if there are any obvious errors in the responses, such as mentioning objects not present in the image. In the case of error, the score is 0; otherwise, it is 1.

\section{Baseline and DFL Evaluations} 
\label{sec:baseline}

This section first presents our evaluation process along with the baseline results. 


The baseline evaluation refers to training a model in a single machine with adequate resources without relying on distributed training. This assessment offers the optimal set of hyperparameters and the number of epochs for each model. Thus, we can assess how closely the performance of DFL deployments matches with the baseline one using the same hyperparameters. Specifically, we use 80\%, 10\%, and 10\% split as training, validation, and test, respectively.
The list of hyperparameters is presented in Table~\ref{table:baseline}. We report the learning rate (LR), batch size, number of epochs to converge, and total epochs for each model. The trained model is tested to check their performance in terms of F1 score except for the MiniGPT4. The F1 score of the SVM and Regression model is 0.95 and 0.94, respectively. ReNet18 and DistilBert offer an F1 score of 0.99 and 0.97, respectively. In the case of MiniGPT4, we usually measure its accuracy and loss as we must ask the minigpt4 questions and check the correctness of the response. Its accuracy is 0.90. 

\begin{table}[]  
\centering  
\caption{ Baseline trained models and their Hyperparameters. } 
\label{table:baseline} 
\adjustbox{max width=\textwidth}{ 
\begin{tabular}{|l|l|l|} 
\hline 
Models & Hyperparameters & \makecell[l]{coverage/total} \\ \hline  
SVM & \makecell[l]{lr: 0.00001 \\ batch size: 1\strut} & 60/500 \\ \hline  
\makecell[l]{Logistic \\ Regression} & \makecell[l]{lr: 0.00001 \\ batch size: 1\strut} & 230/1000 \\ \hline  
ReNet18 & \makecell[l]{lr: 0.001 \\ batch size: 64\strut} & 16/100 \\ \hline  
DistilBert & \makecell[l]{lr: 0.00001 \\ batch size: 16\strut} & 3/20 \\ \hline  
MiniGPT-4 & \makecell[l]{Initial lr: 3e-5 \\ Minimum lr: 1e-5 \\ Warmup lr: 1e-6 \\ Weight decay: 0.05 \\ Iterating: 200 times \\ Warmup step count: 200 \\ batch size: 12 \\ Image size: 224\arraybackslash} & 45/150 \\ \hline  
\end{tabular}  
}  
\end{table}

\textbf{DFL Evaluation:}
In this evaluation, we consider five identical devices (we start the training with the first device) to train a model collaboratively using the same set of hyperparameters used in the baseline evaluation. Thus, the total number of epochs in all DFL configurations is consistent with the baseline. In linear topologies, the number of epochs in each client is equal to the total number of epochs divided by the number of clients. For example, SVM training takes 500 epochs in the baseline; thus, all the participating devices must use at most 500 epochs to complete the training together and be consistent with the baseline. We allocate an equal number of epochs to each device, e.g., 100 epochs for each of the five participating devices when using SVM. In ring topologies, this number is expressed as $epoch = \frac{total\_epoch} {Num\_client * num\_Round}$ because the training is conducted over multiple rounds. For instance, each device is assigned 50 epochs in SVM to be consistent with the total of 500 epochs. Finally, star and mesh topologies use concurrent training; thus, the assigned number of epochs to each device is equal to the total number of epochs divided by the number of rounds. Thus, each device takes 100 epochs as the number of rounds is five. This fixed-epoch strategy helps maintain consistency and fairness across different experiments. By standardizing the number of epochs, we can more accurately compare the performance of different DFL structures under the same conditions. We measure the F1 score as the performance metric in the above setup and evaluations; in particular, in linear and ring topologies, it is the average across five devices, whereas it is the average of multiple rounds in star and mesh topologies. In the following sections, we present the impact of different topologies and the degree of non-IId data on the performance of the five chosen models.

\section{Evaluation Results}\label{sec:evaluation result}

This section presents the results of all models with six different topological deployments of DFL. We present evaluation results with IID and non-IID data and the final takeaway messages. To better understand and avoid confusion, we need to define two concepts: \textit{Transmission Round} for ring topology and \textit{Aggregation Round} for star topology. A \textit{transmission round} refers to the process where all devices sequentially pass the model parameters and complete one training cycle. An \textit{aggregation round} refers to the process where the central server and all clients complete one round of parameter aggregation.

\subsection{Impact of Topologies on the Convergence Rate}
\label{Convergence Evaluation}

The impact of training strategy in linear and ring topologies is different. In the former one, each client only trains a model once irrespective of the training strategy (continuous or aggregated). Thus, the number of epochs in each client is equal to the total number of epochs over the number of clients. On the other hand, the ring topology has two rounds of training; thus, each device gets more than one training chance. Finally, the star and mesh topologies involve concurrent training. In the following, we present the convergence rate of these topologies with IID data.


\textbf{Traditional models:} 
Table~\ref{table: Svm experiment1 Coverage} presents the convergence of SVM with 500 training epochs, following the baseline performance. The model converges after 83 and 92 epochs on the first device in the continuous and aggregate linear topologies, respectively. Consequently, subsequent devices converge as they inherit the converged model from the previous device and use their own IID data.
In the case of ring topology, the model does not converge in the first round. It is observed in the baseline that SVM needs 60 epochs to converge, whereas the first device in the first round gets 50 epochs. Thus, it needs additional epochs during the second round while receiving parameters from the last device of the previous round. Finally, since each device operates concurrently, they exhibit similar convergence behavior in star and mesh topologies.

The convergence behaviour of logistic regression with 1000 epochs is presented in Table~\ref{table: Logical experiment1 coverage}. It exhibits similar performance as in SVM. The model converges in the second device in linear topology, as it needs 230 epochs to converge. However, ring topology resolves this issue, as the first device converges in the second round. Finally, the convergence behaviour of star and mesh topologies is similar to SVM with a higher number of epochs, as the model needs that to converge.

\begin{table}[h]
\centering
\caption{Convergence rate of SVM.}
\label{table: Svm experiment1 Coverage}
\begin{tabular}{lccccc}
\toprule
\textbf{Topology} & \textbf{Client1} & \textbf{Client2} & \textbf{Client3} & \textbf{Client4} & \textbf{Client5} \\
\midrule
C\_linear & 83  & 100 & 200 & 300 & 400 \\
C\_ring   & 250 & 80  & 100 & 150 & 200 \\
A\_linear & 92  & 100 & 200 & 300 & 400 \\
A\_ring   & 250 & 95  & 100 & 150 & 200 \\
Star      & 97  & 99  & 93  & 94  & 95  \\
Mesh      & 96  & 91  & 92  & 98  & 99  \\
\bottomrule
\end{tabular}
\end{table}

\begin{table}[t]
\centering
\caption{Convergence rate of Logical Regression.}
\label{table: Logical experiment1 coverage}
\begin{tabular}{lccccc}
\toprule
\textbf{Topology} & \textbf{Client1} & \textbf{Client2} & \textbf{Client3} & \textbf{Client4} & \textbf{Client5} \\
\midrule
C\_linear & NC  & 230 & 400 & 600 & 800 \\
C\_ring   & 500 & 233 & 300 & 400 & 500 \\
A\_linear & NC  & 250 & 400 & 600 & 800 \\
A\_ring   & 500 & 250 & 300 & 400 & 500 \\
Star      & 400 & 401 & 403 & 400 & 402 \\
Mesh      & 400 & 401 & 402 & 400 & 402 \\
\bottomrule
\end{tabular}
\end{table}


\textbf{Deep neural network models:}
Similar to traditional models, we fixed the total number of epochs in deep learning, keeping it consistent with the baseline. Table~\ref{table: Resnet experiment-coverage 1} and Table~\ref{table: nlp experiment1 coverage} present the convergence rate of ResNet and DistilBERT, where the trend is similar to the traditional models. 
Furthermore, we observe that, since ResNet and DistilBERT are not convex models, unlike traditional models, each device typically has its unique convergence time. This is because when parameters are passed from one device to the next, the loss exhibits fluctuations over 1 to 5 epochs. We speculate that this could be due to the deep structure of models like ResNet, where residual connections introduce non-linear updates during training on different devices, causing such fluctuations. Additionally, we observed that the star and mesh topologies exhibited similar performance to traditional models, as they involve concurrent training. Their results are more stable and do not show fluctuations in loss over several epochs.

\begin{table}[t]
\centering
\caption{ Convergence rate of ResNet. }
\label{table: Resnet experiment-coverage 1}
\begin{tabular}{lccccc}
\toprule
\textbf{Topology} & \textbf{Client1} & \textbf{Client2} & \textbf{Client3} & \textbf{Client4} & \textbf{Client5} \\
\midrule
C\_linear & 18 & 27 & 57 & 75 & 87 \\
C\_ring   & 55 & 17 & 25 & 35 & 46\\
A\_linear & 19 & 25 & 55 & 78 & 90 \\
A\_ring   & 56 & 18 & 25 & 36 & 47 \\
Star      & 16 & 25 & 53 & 77 & 92 \\
Mesh      & 16 & 25 & 53 & 77 & 92 \\
\bottomrule
\end{tabular}
\end{table}

\begin{table}[t]
\centering
\caption{ Convergence rate of DistilBERT. }
\label{table: nlp experiment1 coverage}
\begin{tabular}{lccccc}
\toprule
\textbf{Topology} & \textbf{Client1} & \textbf{Client2} & \textbf{Client3} & \textbf{Client4} & \textbf{Client5} \\
\midrule
C\_linear & 4  & 8  & 11 & 16 & 19 \\
C\_ring   & 12 & 4  & 16 & 18 & 20 \\
A\_linear & 3  & 7  & 12 & 15 & 20 \\
A\_ring   & 11 & 4  & 15 & 18 & 20 \\
Star      & 4  & 8  & 11 & 15 & 19 \\
Mesh      & 4  & 8  & 11 & 15 & 19 \\
\bottomrule
\end{tabular}
\end{table}

\begin{table}[t]
\centering
\caption{Convergence rate of MiniGPT.}
\label{table: LLM experiment1 coverage results}
\begin{tabular}{lccccc}
\toprule
\textbf{Topology} & \textbf{Client1} & \textbf{Client2} & \textbf{Client3} & \textbf{Client4} & \textbf{Client5} \\
\midrule
C\_linear & NC & 45 & 62 & 93 & 123\\
C\_ring   & 65 & 26 & 30 & 45 & 60 \\
A\_linear & NC & 39 & 63 & 91 & 121 \\
A\_ring   & 65 & 24 & 30 & 45 & 60 \\
Star      & 40 & 40 & 41 & 40 & 42 \\
Mesh      & 40 & 41 & 40 & 41 & 40 \\
\bottomrule
\end{tabular}
\end{table}

\textbf{Large language models:}
Table~\ref{table: LLM experiment1 coverage results} presents the convergence rate of MiniGPT. Like deep learning models, MiniGPT loss exhibits fluctuations when parameters are passed from one device to another, as the model is not convex. The behaviour across topologies is expected and aligned with the previous trend. Specifically, the model converges consistently over star and mesh topologies.   

\begin{table}[t]
    \centering
    \caption{The average F1 score on different DFL deployments.}
    \label{table:F1Score}
    \adjustbox{max width=\textwidth}{%
    \begin{tabular}{|p{1.2cm}|p{0.9cm}|p{0.8cm}|p{0.9cm}|p{0.8cm}|p{0.6cm}|p{0.6cm}|}
        \hline
        \textbf{Model} & \textbf{C\_linear} & \textbf{C\_ring} & \textbf{A\_linear} & \textbf{A\_ring} & \textbf{Star} & \textbf{Mesh} \\ \hline
        SVM         & 0.964 & 0.960 & 0.957 & 0.957 & 0.959 & 0.960  \\ \hline
        Logistic    & 0.938 & 0.946 & 0.937 & 0.940  & 0.944 & 0.944 \\ \hline
        ResNet      & 0.987 & 0.979 & 0.985 & 0.981 & 0.985 & 0.969 \\ \hline
        DistilBERT  & 0.938 & 0.946  & 0.937 & 0.940 & 0.944 & 0.944\\ \hline
    \end{tabular}%
    }
\end{table}

\textbf{F1 Score:} We present the average F1 score of all models over all topologies in Table~\ref{table:F1Score}. The results indicate that once a model converges, it offers a good score proportional to its convergence rate.  
Note that we can only measure the accuracy of MiniGPT, which is 0.9 in C\_linear, star, and mesh topologies and 0.85 in C\_ring, A\_linear, and A\_ring topologies. Overall, once a model converges in a topology, its performance matches the baseline. The comparison shows that the F1 scores under different topologies are almost identical to the baseline. This indicates that the impact of various topologies on model performance is minimal, and the effectiveness of different deployment methods remains largely consistent. Specifically, the F1 scores of SVM and Logistic Regression under different topologies are almost identical to the baseline, with maximum differences of $±0.014$ and $±0.006$, respectively. ResNet shows a maximum difference of $0.021$, while DistilBERT exhibits a maximum difference of $0.033$ within acceptable limits. Furthermore, MiniGPT-4’s baseline accuracy is 0.9, while Experiment 1 achieves an accuracy of either 0.9 or 0.85, also very close to the baseline. These results suggest that the combinations of different models and topologies have a negligible effect on performance, maintaining a high level of consistency.

\begin{boxH}
\textbf{Key Takeaways:}
We observe that whether the local model is a traditional model, a deep learning model, or an LLM, it successfully converges, and the performance matches the baseline with IID data. However, different topologies require varying computations (epochs), e.g., star and mesh offer the most stable training due to their concurrent training nature. Linear and ring, on the other hand, require adequate computation in each device for model convergence. Finally, deep neural network models may suffer from fluctuations in loss functions while parameters move from one device to another due to their non-convex nature.  
\end{boxH}

\subsection{Impact of Non-IID data Distribution}

This evaluation aims to investigate the impact of the degree of non-IID data on the performance of DFL with the same hyperparameters as in the baseline model. We evaluate multiple degrees of nonIID data as stated in Section~\ref{sec:implementation}, specifically \([0.5,0.6,0.7,0.8,0.9]\), \([0.1,0.3,0.5,0.7,0.9]\), and \([1,0,0.7,1,0]\). We discuss the evaluations results from all three distributions while include only the Level 2 non-IID distribution (\([0.1,0.3,0.5,0.7,0.9]\)) evaluation outcome due to space constraints. 
Also, we do not evaluate the performance of MiniGPT-4 on non-IID data because its data is inherently non-IID, i.e., different images are associated with different texts with a non-IID distribution.

\textbf{Traditional models:} 
Table~\ref{table:Svm_experiment2_coverage_level2} presents the convergence rate of the SVM with the Level 2 distribution. In the case of linear and ring topologies, irrespective of their training strategies, the model converges only in two devices. We suspect that this is because the data in these two devices have a better label balance than other devices. However, the linear and ring topologies cannot solve such imbalance issues due to the nature of their training. The model, however, converges in all devices over star and mesh topologies, i.e., they offer the best performance as in the case of IID distribution. Concurrent training seems beneficial over sequential training in DFL deployments with non-IID data distribution, where participating devices learn from others.  
Table ~\ref{table:Logic_experiment2_coverage_level2} presents the performance of the logistic regression model with level 2 non-IID data. The performance trend in linear and ring topologies is similar to that of the SVM model, i.e., the model converges on two devices with a better label distribution compared to the other devices. Similarly, star and mesh topologies offer the best performance due to their concurrent communications. The performance trend in Level 1 distribution is similar to that of Level 2 due to the same reasons. However, Level 3 results exhibit an interesting but expected performance trend, where the trend just reversed between linear and ring vs. star and mesh. The first two devices in linear and ring topologies complement each other with their labelled data, which is carried over to the remaining devices. However, star and mesh experience an oscillation in this distribution and suffer from performance degradation. 



\begin{table}[!t]
    \centering
    \caption{SVM convergence rate with non-IID data.}
    \label{table:Svm_experiment2_coverage_level2}
    \begin{tabular}{lccccc}
        \toprule
        \textbf{Topology} & \textbf{Client1} & \textbf{Client2} & \textbf{Client3} & \textbf{Client4} & \textbf{Client5} \\
        \midrule
        C\_linear & NC  & NC  & 260 & 301 & NC  \\
        C\_ring   & NC  & NC  & 400 & 450 & NC  \\
        A\_linear & NC  & NC  & 299 & 302 & NC  \\
        A\_ring   & NC  & NC  & 398 & 452 & NC  \\
        Star      & 402 & 400 & 401 & 400 & 402 \\
        Mesh      & 403 & 400 & 401 & 400 & 401 \\
        \bottomrule
    \end{tabular}
\end{table}

\begin{table}[htbp]
    \centering
    \caption{Logic Regression convergence rate with non-IID data.}
    \label{table:Logic_experiment2_coverage_level2}
    \begin{tabular}{lccccc}
        \toprule
        \textbf{Topology} & \textbf{Client1} & \textbf{Client2} & \textbf{Client3} & \textbf{Client4} & \textbf{Client5} \\
        \midrule
        C\_linear & NC  & NC  & 599 & 799 & NC  \\
        C\_ring   & NC  & NC  & 799 & 845 & NC  \\
        A\_linear & NC  & NC  & 599 & 799 & NC  \\
        A\_ring   & NC  & NC  & 799 & 845 & NC  \\
        Star      & 851 & 850 & 852 & 851 & 849 \\
        Mesh      & 830 & 831 & 832 & 831 & 831 \\
        \bottomrule
    \end{tabular}
\end{table}

\textbf{Deep neural network models:} 
Table ~\ref{table:Resnet_experiment2_coverage_level2} and Table ~\ref{table:nlp_experiment2_coverage_level2} present the convergence rate of ResNet and DistilBERT over Level 2 non-IID data distribution, respectively. 
For Resnet, we observe that at Non-IID level 1, all DFL models can converge. This indicates that at level 1, various topologies perform well. In both cases, star and mesh topologies offer the best performance, as we have seen in the case of traditional models. Similarly, the performance of linear and ring suffers. Specifically, devices with better-balanced labels converge. Level 1 also offers a similar performance trend, while level 3 distribution is the worst in terms of performance across all topologies. We found that in DFLs with non-star or non-mesh topologies, convergence is often complex when the data distribution on the devices is highly uneven. In other words, star and mesh are the most suitable deployment strategies to host deep learning models in DFL deployments. 

\begin{table}[htbp]
    \centering
    \caption{ Convergence rate of ResNet on non-IID data. }
    \label{table:Resnet_experiment2_coverage_level2}
    \begin{tabular}{lccccc}
        \toprule
        \textbf{Topology} & \textbf{Client1} & \textbf{Client2} & \textbf{Client3} & \textbf{Client4} & \textbf{Client5} \\
        \midrule
        C\_linear & NC & 39 & 59 & NC & NC \\
        C\_ring   & NC & 20 & 30 & 40 & NC \\
        A\_linear & 19 & 39 & 59 & 79 & 99 \\
        A\_ring   & NC & 20 & 30 & 40 & NC \\
        Star      & 24 & 24 & 24 & 24 & 24 \\
        Mesh      & 24 & 24 & 24 & 24 & 24 \\
        \bottomrule
    \end{tabular}
\end{table}

\begin{table}[htbp]
    \centering
    \caption{ Convergence rate of DistilBERT on non-IID data. }
    \label{table:nlp_experiment2_coverage_level2}
    \begin{tabular}{lccccc}
        \toprule
        \textbf{Topology} & \textbf{Client1} & \textbf{Client2} & \textbf{Client3} & \textbf{Client4} & \textbf{Client5} \\
        \midrule
        C\_linear & NC & 8  & 12 & 16 & 20 \\
        C\_ring   & NC & 12 & 14 & 18 & NC \\
        A\_linear & NC & NC & 12 & 16 & 20 \\
        A\_ring   & NC & 12 & 14 & 18 & NC \\
        Star      & 20 & 20 & 20 & 20 & 20 \\
        Mesh      & 20 & 20 & 20 & 20 & 20 \\
        \bottomrule
    \end{tabular}
\end{table}


\textbf{F1 Score:}
The average F1 score of different DFL deployments for the Level 2 non-IID data distribution is shown in Figure~\ref{table:F1Score-nonIID}. The results indicate that the F1 score decreases when the level of non-IID data distribution increases. Also, their F1 score is proportional to the convergence rate once the models converge. The impact of non-IID data on different topologies for model performance is minimal, indicating that these deployments exhibit good stability and consistency on non-IID data. Specifically, the maximum F1 score difference for the SVM model is 0.029, with the best performance in the C\_ring topology (0.825) and the lowest in the A\_linear topology (0.796). The maximum difference for Logistic regression is 0.099, performing best in the Mesh topology (0.834) but worst in the C\_linear and A\_linear topologies (0.742 and 0.735, respectively). ResNet shows the minimal F1 score variation, with a maximum difference of 0.021, demonstrating stable performance. However, DistilBERT has a maximum difference of 0.109, with performance fluctuations, particularly in the star and mesh topologies. ResNet performs best, while SVM and logistic regression models show minimal fluctuation, with DistilBERT offering the worst.


\begin{table}[htbp]
    \centering
    \caption{The average F1 score of different DFL deployments for non-IID data.}
    \label{table:F1Score-nonIID}
    \adjustbox{max width=\textwidth}{%
    \begin{tabular}{|p{1.2cm}|p{0.9cm}|p{0.8cm}|p{0.9cm}|p{0.8cm}|p{0.6cm}|p{0.6cm}|}
        \hline
        \textbf{Model} & \textbf{C\_linear} & \textbf{C\_ring} & \textbf{A\_linear} & \textbf{A\_ring} & \textbf{Star} & \textbf{Mesh} \\ \hline
        SVM         &0.813 & 0.825 &0.796 & 0.809 & 0.806 & 0.798  \\ \hline
        Logistic    & 0.742 & 0.791 & 0.735 & 0.783  & 0.783 & 0.834 \\ \hline
        ResNet      & 0.967 & 0.954 & 0.975 & 0.967 & 0.974 & 0.963 \\ \hline
        DistilBERT  & 0.818 & 0.744  & 0.834 & 0.743 & 0.730 & 0.725\\ \hline
    \end{tabular}
    }
\end{table}

\begin{boxH}
\textbf{Key Takeaways:}
If data is evenly distributed on each device, traditional and deep learning algorithms better converge with high F1 scores. We also observe that the impact of network topology on device convergence is more dominant than the aggregation strategies. In the traditional model, factors like the initiating device and data distribution both have an impact on the convergence. In particular, if devices can chronologically learn from preceding neighbours, they exhibit better performance in the linear and ring. On the contrary, stars and rings perform well if devices can concurrently train their datasets and aid others. Both traditional and deep learning models converge with high performance with evenly distributed data on each device. Specifically, deep learning models perform poorly with a high degree of non-IID data. This suggests that devices with incomplete label datasets in real-world applications should be excluded, as they negatively impact the convergence of the global model. When devices have complete label sets, star and mesh topologies perform better with uneven label distributions. 
\end{boxH}
\section{Conclusion}
\label{section: Conclusion}

This work investigated the impact of network topology, non-IID data, and training strategies on the convergence of DFL. Specifically, we conducted both mathematical analysis and experimental evaluations to study these factors comprehensively.
Using convex optimization, we analyzed the convergence of six different DFL deployments. We found that when data distribution across devices is IID, the difference between the ideal and actual solutions is constant. As the degree of non-IID data increases, this difference becomes larger. Next, we conducted two experiments to analyze the convergence rates of traditional, deep learning, and large language models in six different DFL deployments. The evaluation results aligned with the mathematical analysis, i.e., all models converge on each DFL deployment with IID data. However, in the case of non-IID data, the convergence rate is inversely proportional to the degree of non-IID data. In the future, we plan to leverage software-defined networking to realize agile DFL deployments.

\ifCLASSOPTIONcaptionsoff
  \newpage
\fi
\bibliographystyle{IEEEtran}
\bibliography{reference}

\end{document}